\documentclass[letterpaper, 10 pt, conference]{ieeeconf}  % Comment this line out if you need a4paper

\IEEEoverridecommandlockouts                              % This command is only needed if 
\usepackage{algorithm}
\usepackage{algcompatible}
\usepackage{algpseudocode}
\usepackage{mathtools}

\makeatletter
\let\OldStatex\Statex
\renewcommand{\Statex}[1][3]{%
  \setlength\@tempdima{\algorithmicindent}%
  \OldStatex\hskip\dimexpr#1\@tempdima\relax}
\makeatother
\usepackage{graphics} % for pdf, bitmapped graphics files
\usepackage{url}
\usepackage{booktabs}

\usepackage{amsmath} % assumes amsmath package installed
\usepackage{amsfonts}
\usepackage{amssymb}  % assumes amsmath package installed
\usepackage{multicol}
\usepackage{graphicx}
\usepackage{svg}
\usepackage{layouts}
\usepackage{wrapfig}
\usepackage{caption}
\usepackage{graphicx,subcaption}
\usepackage{easyReview}
\usepackage{printlen}
\usepackage{bm}
\usepackage{siunitx}
\usepackage{colortbl}	% to color table background
\usepackage[utf8]{inputenc}
\usepackage{pgf}
\usepackage{tikz}
\usepackage{tikzscale}
\usepackage{pgfplots}
\DeclareUnicodeCharacter{2212}{\textendash}
\pgfplotsset{compat=1.14}
\usepgfplotslibrary{external}
\usepgfplotslibrary[external]
\tikzset{external/force remake}
\usepackage{algorithm}
\usepackage{algpseudocode}
\usepackage{makecell}
\usepackage{hyperref}

\title{\LARGE \bf
Contact-conditioned learning of multi-gait locomotion policies
}

\author{Michal Ciebielski$^{1}$, Federico Burgio$^{2}$, Majid Khadiv$^{1}$\thanks{%
$^{1}$ Munich Institute of Robotics and Machine Intelligence, Technical University of Munich, Germany. \texttt{\small michal.ciebielski@tum.de, majid.khadiv@tum.de}\newline \indent
$^{2}$ Department of Industrial Engineering, University of Trento, Italy. \texttt{\small federico.burgio@studenti.unitn.it}}}

\begin{document}

\maketitle
\thispagestyle{empty}
\pagestyle{empty}

% \textcolor{red}{\textbf{TODO:} Images of robot hopping here (sim + real?).}

%%%%%%%%%%%%%%%%%%%%%%%%%%%%%%%%%%%%%%%%%%%%%%%%%%%%%%%%%%%%%%%%%%%%%%%%%%%%%%%%
\begin{abstract}
In this paper, we examine the effects of goal representation on the performance and generalization in multi-gait policy learning settings for legged robots. To study this problem in isolation, we cast the policy learning problem as imitating model predictive controllers that can generate multiple gaits. We hypothesize that conditioning a learned policy on future contact switches is a suitable goal representation for learning a single policy that can generate a variety of gaits. Our rationale is that policies conditioned on contact information can leverage the shared structure between different gaits. Our extensive simulation results demonstrate the validity of our hypothesis for learning multiple gaits on a bipedal and a quadrupedal robot. Most interestingly, our results show that contact-conditioned policies generalize much better than other common goal representations in the literature, when the robot is tested outside the distribution of the training data.
\end{abstract}

%%%%%%%%%%%%%%%%%%%%%%%%%%%%%%%%%%%%%%%%%%%%%%%%%%%%%%%%%%%%%%%%%%%%%%%%%%%%%%%%
\section{Introduction}

The dominant approaches to controlling legged robots are model-based control, in particular (nonlinear)
model predictive control (MPC) \cite{meduri2023biconmp,sleiman2023versatile,mastalli2023agile}, and learning-based control, especially (deep) reinforcement learning (RL) \cite{tan2018sim,hwangbo2019learning,xie2020learning,siekmann2021blind}. Each of these approaches presents different strengths and weaknesses. MPC techniques are notably versatile, meaning that given adequate perception and task information, they can easily adapt to new tasks and environments. Furthermore, they can take the constraints of the robot and environment into account to generate safe behaviors. However, they are sensitive to unexpected contact events and tend to be computationally expensive at run-time. These difficulties have caused a major shift towards learning-based policy optimization techniques and sim-to-real transfer. By directly optimizing the parameters of a feedback policy (usually a neural network), the run-time computation in these approaches is at a minimum. Furthermore, by randomizing the uncertain aspects of the problem during training using domain randomization \cite{tobin2017domain}, the resulting policy is robust and can be directly transferred to real robots.

Despite these advances, learning \emph{multi-gait} control policies remains a challenging problem. Notable efforts to generate such policies include \cite{huang2024diffuseloco}, which leverages an offline dataset of demonstrations to enable a single, skill-conditioned generative policy to perform five different gaits. In \cite{reske2021imitation}, a quadruped robot learns multiple gaits through a mixture-of-experts policy network, where each expert specializes in a single gait. The interesting feature of this framework is that it can leverage the shared structure between the gaits through a shared part of the neural network among all gaits. However, both of these approaches are constrained by a limited number of gaits and they cannot scale to the case of generating a wide range of cyclic and acyclic behaviors.

One key aspect when learning multiple skills is to have a goal representation that not only encompasses all possible locomotion behaviors, but it also encodes the shared structure between different behaviors. However, the common choice in the literature is to condition locomotion policies on an average horizontal velocity that fails to address both of these aspects. Firstly, a robot can achieve similar average velocities with many different cyclic or acyclic behaviors. Hence, conditioning on the desired velocity does not distinguish between different ways a robot can move. Secondly, conditioning on the desired velocity (and skill) fails in capturing the similarities between different skills.

Since making and breaking contact is essential in realizing any interesting locomotion behavior, we hypothesize that a goal representation based on the properties of contact switches can be an excellent candidate.

Goal representations based on contact have been used in policy learning in \cite{peng2017deeploco} and \cite{zhang2024wococo}. However, in both works, this information is used to learn single skill policies. In contrast, we hypothesize that the contact goal representation is particularly useful in case of multi-gait learning. 

Hence, in this paper, we present contact-conditioned policies that can be used with any high-level contact planners and benchmark their performance against commonly used goal-conditioning based on velocity and gait. Specifically, we train multi-gait contact-conditioned policies via behavior cloning on an expert MPC. 
The main contributions of the paper are as follows: 

\begin{itemize}
    \item We propose a general goal representation based on contact information that in principle can learn any cyclic and acyclic behaviors and can be coupled with any contact planner. 
    % ***This representation is general enough that can be used for learning any type of gaited or non-gaited locomotion behaviors through a single policy network.\textbf{(last time in the feedback they said we cannot justify non-gaited. Now we have some examples of stopping behavior so we could somehow argue that. otherwise we drop the claim of non-gaited)}
    \item We benchmark the performance and robustness of a multi-gait contact-conditioned policy against gait and velocity-conditioned policies typically found in the literature. Our systematic comparison on two different legged systems, a biped and a quadruped, shows that policies conditioned on contact information consistently outperform those based on velocity and gait, both within and outside the training distribution. Overall, we show that not only does the contact-conditioned policy enhance single-gait performance, it also boosts multi-gait learning, achieves the best results in our most challenging tests and exhibits superior out-of-distribution generalization.
\end{itemize}

% Through a systematic comparison, we show that conditioning the low-level policy on contact not only improves robustness and performance for a single gait, but also improves the multi-gait learning. 

The remainder of the paper is structured as follows: Section \ref{sec:preliminary} overviews the required fundamentals to start with the new approach. Section \ref{sec:method} gives a detailed description of the proposed approach and implementation details. Section \ref{sec:result} presents the results on the comparison between the contact mode goal representation and the gait and velocity goal representation. 
% In Section \ref{sec:discussion}, we discuss the obtained results. 
Finally, Section \ref{sec:conclusion} summarizes our findings and proposes several directions for future works.

%The reason why we use MPC as an expert: Each of these approaches presents different strengths and weaknesses. MPC techniques are notably versatile, meaning that given adequate perception and task information, they can easily adapt to new tasks and environments. Furthermore, they can take the constraints of the robot and environment into account to generate safe behaviors. However, they are sensitive to unexpected contact events and tend to be computationally expensive at run-time. 

\section{Preliminaries}\label{sec:preliminary}
In this section, we give an overview of the preliminaries which are necessary to perform behavior cloning from MPC. The reason behind this choice is to focus on the effects of policy representation, without dealing with exploration nuances in DRL frameworks. To make the paper self-contained, in the following we explain the framework used in this paper. However, our results are independent of the choice of the expert controller.

We first describe contact explicit MPC, which is the expert demonstrator in our framework. Then we describe how to generate a diverse set of initial conditions for a given contact sequence.  Finally, we present the general behavior cloning from MPC algorithm. %Note that our results are independent of these frameworks and should apply to any other approach to learning locomotion policies.

% \subsection{Behavioral cloning}
% Behavioral cloning (BC) is the simplest form of imitation learning through which the collected data from an expert is used to train a policy using supervised learning \cite{pomerleau1988alvinn}. While several extensions of online algorithms have been proposed to eliminate the compounding error and distribution shift \cite{ross2010efficient,levine2013guided}, BC is still used for many applications due to its simplicity and ease of implementation. The main reason behind the use of BC in this work is that, contrary to online learning and RL algorithms, the distribution of data remains the same for our evaluation which allows us to focus mainly on the choice of goal representation.

% In BC process, first, a dataset $D$ comprised of observed states $s$ and expert actions $a$  is collected offline. Then, these samples are treated as if they were independent and identically distributed (i.i.d) and supervised learning is carried out to approximate a policy $\pi$ that replicates the action of the expert at any given state. The training objective is minimizing the errors in mimicking an expert's actions $\pi = \arg\max_{\pi} \mathbb{E}_D[\log \pi(a \mid s)]$, simply minimizing the empirical risk as in supervised learning. 
\subsection{Contact Explicit MPC}
MPC solves a receding horizon optimal control problem by optimizing a sequence of future control actions subject to dynamics and path constraints. In contact-explicit formulation, to avoid solving a nonsmooth optimization, the contact sequence is planned first using a given contact planner and then an MPC controller solves a smooth optimal control problem:
\begin{equation}
\label{eq:mpc}
    \begin{aligned}
    \min_{x, u, t_{1:k-1}} & \quad \sum_{t=0}^{N-1} 
    \phi_{t}(x_{t}, u_{t}) + \phi_{T}(x_{T}) \\
    \text{s.t.} \quad & x_{t+1} = f_m(x_t,u_t),\\%, \quad t=0,\dots,N-1,\\
    % & \quad J(q)\ddot{q}+\dot{J}(q,v)\dot{q} = 0 \\
    & h_m(x_t, u_t) \leq 0, \quad \forall t \in [t_{m-1}, t_m), \; m=1,..,k, \\
    & h(x_t, u_t) \leq 0, \quad t = 0, 1 \dots, N-1 \\
    & x_0 = x _{init} \\
    \end{aligned}
\end{equation}
where $x$ and $u$ represent the state and control trajectories. $f_m$ is the dynamics of the system which changes depending on the mode $m=1,..,k$. In addition to global path constraints $h$ such as torque and joint limits, there are also mode dependent constraints $h_m$ such as contact constraints. MPC solves the multi-stage optimal control problem in \eqref{eq:mpc}, starting from the measured state $x_{init}$.

\subsection{Generating Diverse Initial Conditions}
A key ingredient in imitation learning is generating a diverse dataset. When cloning MPC, it is important to consider a diverse distribution of $x_{init}$ from which the MPC is rolled out. To do so, we first generate one rollout of the MPC in simulation and consider that as a nominal trajectory, then perturb the robot around this trajectory and roll out the MPC from the resulting $x_{init}$. However, naively perturbing the initial condition does not respect the contact constraints and may even result in ground penetration which is an infeasible initial condition for \eqref{eq:mpc}. To achieve contact-consistent perturbations, we follow the method in \cite{khadiv2023learning}. Namely, we first sample an unconstrained perturbation from a normal distribution $\delta \mathbf{q}, \delta\mathbf{v} \sim \mathcal{N}(\mu,\sigma^2)$, where $\mathbf{q}$ and $\mathbf{v}$ are the generalized coordinates and velocities of the robot. Then, we project this perturbation in the nullspace of the contact constraints:
\begin{equation}
\label{eq:contact_consistent_perturbations}
    \begin{bmatrix} \delta \mathbf{q}_c \\ \delta \mathbf{v}_c \end{bmatrix} =
    \left( I - A_c^{\dagger} A_c \right)
    \begin{bmatrix} \delta \mathbf{q} \\ \delta \mathbf{v} \end{bmatrix},
\end{equation}
where matrix $A_c$ is constructed as
\begin{equation}
    \label{eq:contact_nullspace}
    A_c = \begin{bmatrix} J_c & 0 \\ \dot{J}_c & J_c \end{bmatrix}.
\end{equation}

In these equations, $I$ is the identity matrix and $J_c$ is the concatenation of the jacobians of the active contact points at the current time. The projected perturbations are added to the initial state of the robot $\mathbf{q}\oplus\delta \mathbf{q}_c$ and $\mathbf{v}+\delta \mathbf{v}_c$ and constitute $x_{init}$ in \eqref{eq:mpc}.

\begin{algorithm}[ht]
\caption{Behavior cloning from MPC}
\label{alg:bcmpc}
\begin{algorithmic}[1]
\State Given initial conditions $\mathcal{X}$, number of rollouts $N$, failed states $\mathcal{S}_{\text{failed}}$ and contact plans $\mathcal{G}_{1:N}$.
\For{$i = 1, \dots, N$}
    \State $s_0 \sim \mathcal{X}$ 
    \Comment{Sample initial condition: eq. \ref{eq:contact_consistent_perturbations}}
    \For{$t = 0, \dots, T$}
        \State $u_t = \arg\min_{s,u} C(s_t, \mathcal{G}_i)$\Comment{Solve optim. \ref{eq:mpc}}
        \State $s_{t+1} = f_{\text{sim}}(s_t, u_t)$ \Comment{Roll out MPC in sim.}
        \If{$s_{t+1} \in \mathcal{S}_{\text{failed}}$}
            \State \textbf{return} infeasible and break
        \EndIf
        \State Add sample: $\mathcal{D}_{s,a}^i \gets \{s_t, a_t\}$
    \EndFor
    \If{feasible}
        \State Add feasible rollout to the dataset: $\mathcal{D}_{s,a} \gets \mathcal{D}_{s,a}^i$
    \EndIf
\EndFor
\State : $\pi_{\theta} = \arg\max_{\pi_{\theta}} \mathbb{E}_{\mathcal{D}_{s,a}} [\log \pi_{\theta}(a | s)]$ \newline
\Comment{Perform supervised learning}
\end{algorithmic}
\end{algorithm}

\subsection{Behavior Cloning from MPC}

To perform behavior cloning from MPC under a fixed contact sequence, Algorithm \ref{alg:bcmpc} is used. First, contact-consistent perturbations are performed to sample feasible initial conditions. Then, the contact-explicit MPC \eqref{eq:mpc} is rolled out in simulation for a given contact plan $\mathcal{G}_i$. If the robot does not fail, the data from the episode is added to a dataset $D$ comprised of observed states $s$ and expert actions $a$. This offline-collected data is treated as if they were independent and identically distributed (i.i.d) and supervised learning is carried out to approximate a policy $\pi$ that replicates the action of the expert at any given state. The training objective is minimizing the errors in mimicking an expert's actions $\pi = \arg\max_{\pi} \mathbb{E}_D[\log \pi(a \mid s)]$, simply minimizing the empirical risk as in supervised learning. 

\textbf{Remark} While one can use some variation of online algorithms such as DAGGER \cite{ross2011reduction} to avoid distribution shift and better sample-complexity \cite{pua2024safe}, here, we intentionally limit ourself to the BC formulation to have a static dataset and study the goal representation learning problem in isolation.

% \subsection{Expert Controllers}\label{sec:expert}
% We use the MPC formulation in \cite{boroujeni2021unified} that extends the formulation in \cite{daneshmand2021variable,khadiv2020walking} to a unified framework for walking and running of bipedal robots. The software is also open-source and available on \href{https://github.com/machines-in-motion/reactive_planners}{Github}. Due to its simplicity of formulation, the MPC can solve for both the step location and timing of the next step at 1 KHz. This makes the MPC highly reactive to external disturbances and changes of commands (desired gait and motion), which are the main requirements for learning a policy in this work. The generated foot trajectories from the MPC are mapped to the joint torques using task-space inverse dynamics. 

% We also use the nonlinear MPC formulation in \cite{meduri2023biconmp} which, given a desired contact sequence and switch timings, uses an iterative kino-dynamic solver to generate whole body trajectories in a receding horizon fashion. It is also open source and available on \href{https://github.com/machines-in-motion/biconvex_mpc}{Github}. We would like to emphasize that our results in this paper are independent of the choice of the MPC and are valid for any other expert controller.

\section{Method}\label{sec:method}
In this section, first we present our proposed conditioning on the next contact modes and switch timings. Then, we outline the details of how we implemented the general BC from MPC algorithm \ref{alg:bcmpc}.

\subsection{Goal Representation}
% \begin{figure}
%     \centering
%     \includegraphics[width=\linewidth]{figures/policy_structure.png}
%     \caption{Velocity-conditioned versus contact-conditioned policy structure}
%     \label{fig:networsk}
% \end{figure}

We define $\mathcal{C}$ as the contact status of all $E$ end-effectors of a legged robot (we omit sliding and only consider breaking and sticking contact):
\begin{equation}
    \mathcal{C} = \left( c_1, c_2, \dots, c_E \right), \quad c_i \in \{0,1\},
\end{equation}
where $c_i=1$ specifies that the $i$th end-effector is in contact with the environment.

\begin{figure}[H]    \centerline{\includegraphics[scale=0.20]{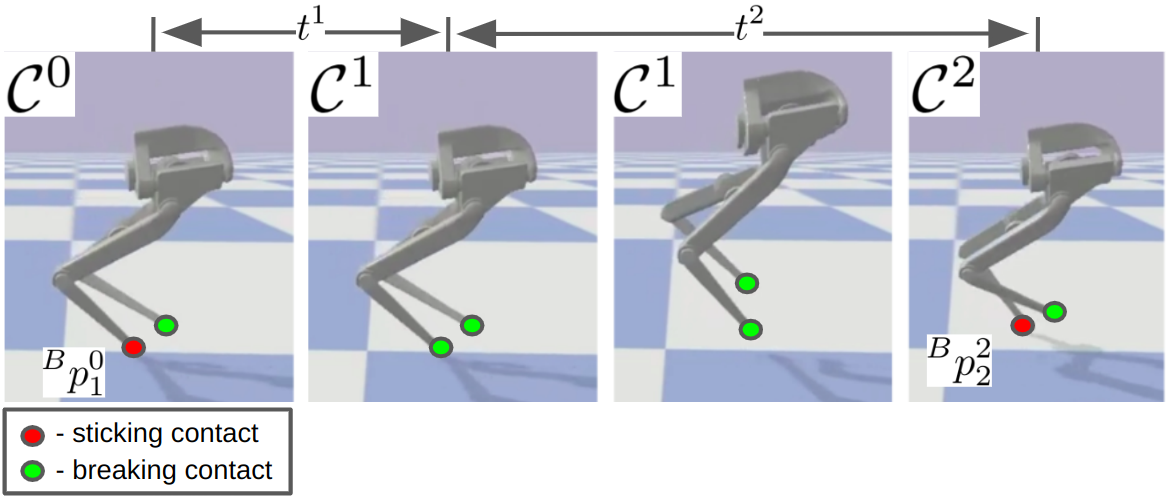}}
    \caption{Example contact sequence captured on the biped Bolt. The left panel represents the current state of the robot and the following three panels depict two contact switches.}
    \label{fig:contact_goal_vis}
\end{figure}
% In more general settings, $c_i$ can take integer values that represent contact with different objects in the environment including fixed and dynamic, however in our setting the robot only interacts with the floor, making $c_i$ binary.
% Define the goal sequence
A contact goal sequence $\mathcal{G}^K$ is defined as a sequence of $K$ desired contact modes, end effector positions and their corresponding switching timings:
\begin{equation}
    \mathcal{G}^K = \{ (\mathcal{C}^1, \prescript{B}{}{p}_{1:N}^1, t^1), \dots, (\mathcal{C}^K, \prescript{B}{}{p}_{1:N}^K, t^K) \},
\end{equation}
where each $\mathcal{C}^k$ is the desired contact mode at the $k$-th switching event, $\prescript{B}{}{p}_i^k$ is the desired position for the $i$-th end-effector in the base frame and $t^k$ is the time at which the switch occurs. For an end effector in breaking contact $\prescript{B}{}{p}_{ee}^k = \mathbf{0}_3$. In our setting it was sufficient to provide information on one mode switch into the future and its corresponding positions and timing, i.e. $\mathcal{G}^1 = (\mathcal{C}^1, \prescript{B}{}{p}_{1:N}^1, t^1)$. However, one could potentially increase the number of contact switches into the future for motions on more complex environments. This representation is general and encompasses both cyclic and acyclic behaviors. Figure \ref{fig:contact_goal_vis} visualizes an example of such representation on the biped robot Bolt.

\begin{figure}[H]
    \centerline{\includegraphics[scale=0.30]{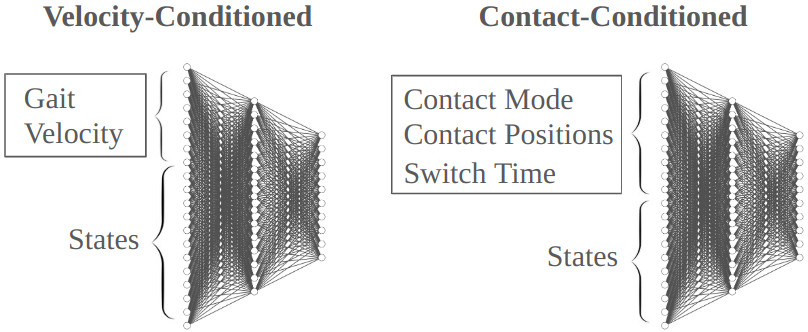}}
    \caption{Policy structure for the velocity-conditioned policy and the contact-conditioned policy.}
    \label{fig:networks}
\end{figure}

We benchmark the contact mode goal representation $g_{con}$ against the standard desired velocity goals $g_{\text{vel}} := [v_{\text{des}}, b, \phi]$. The variables in this representation are the desired velocity $v_{des} \in \mathbb{R}^3$, the type of gait ($b \in \{\text{walk, run\}}$ for the biped, and $b \in \{\text{trot, jump}\}$ for the quadruped) and a phase variable $\phi$ that goes from 0 to 1 for the duration of the gait cycle.
\begin{table}[ht]
    \centering
    \caption{Goal Representations for Trained Policies}
    \label{tab:goal_representations}
    \begin{tabular}{ll}
        \toprule
        \textbf{Policy} & \textbf{Goal Representation} \\
        \midrule
        \emph{Con.}           & $g_{\text{con}} := \mathcal{G}^1$ \\
        \emph{Vel.+Gait}           & $g_{\text{vel}} := [v_{\text{des}}, b, \phi]$ \\
        \emph{Vel.+Gait+Con.}      & $g_{\text{vel+con}} := [v_{\text{des}}, b, \phi, \mathcal{G}^1]$ \\
        \emph{Con.+Gait}      & $g_{\text{con+gait}} := [\mathcal{G}^1, b]$ \\
        \bottomrule
    \end{tabular}
\end{table}
In addition to comparing the contact-conditioned policy \emph{Con.} and the velocity-conditioned policy \emph{Vel.+Gait}, we also perform two ablations.   \emph{Vel.+Gait+Con.} combines $g_{vel}$ and $g_{con}$ and \emph{Con.+Gait} combines the gait type $b$ to the contact mode goal $g_{con}$. All considered goal representations are summarized in Table \ref{tab:goal_representations}. A visual depiction of the policy structure of the velocity and contact-conditioned policies is shown in Fig. \ref{fig:networks}.

% \subsection{Policy}
\subsection{Implementation}
In this section, we give an overview of our implementation specifications.

\subsubsection{Expert controllers}

To control the biped Bolt (Fig. \ref{fig:contact_goal_vis}) we used the MPC formulation in \cite{boroujeni2021unified} that extends the formulation in \cite{daneshmand2021variable,khadiv2020walking} to a unified framework for walking and running of bipedal robots. The software is also open-source and available on \href{https://github.com/machines-in-motion/reactive_planners}{Github}. Due to its simplicity of formulation, the MPC can solve for both the step location and timing of the next step at 1 KHz. This makes the MPC highly reactive to external disturbances and changes of commands (desired gait and motion), which are the main requirements for learning a policy in this work.
% The generated foot trajectories from the MPC are mapped to the joint torques using task-space inverse dynamics. 

To control the quadruped Go2 we used the MPC formulation in \cite{meduri2023biconmp} which, given a desired contact sequence and switch timings, uses an iterative kino-dynamic solver to generate whole body trajectories in a receding horizon fashion. It is also open-source and available on \href{https://github.com/machines-in-motion/biconvex_mpc}{Github}. We would like to again emphasize that our results in this paper are independent of the choice of the MPC and are valid for any other expert controller.
\subsubsection{Data Collection} %500k samples
Data for the behavior cloning was collected by rolling out simulations for both the biped and quadruped with the expert controllers. One rollout of the simulation is defined as a sequence of desired velocities, durations, and gaits which are sampled for both robots with the values shown in Table \ref{tab:rollout_parameters}. The velocities $v_x$, $v_y$ and time durations $T_{dur}$ are sampled from a uniform distribution $\mathcal{U}$. The gaits are sampled from an equal categorical distribution.

\begin{table}[h]
    \centering
    \renewcommand{\arraystretch}{0.3}
    \setlength{\tabcolsep}{4pt}
    \caption{Rollout parameter definitions.}
    \begin{tabular}{l c c c c}
        \toprule
        Robot & \( v_x \) & \( v_y \) & \( T_{dur} \) & Gaits $b$ \\
        \midrule
        Biped & \( \mathcal{U}(-1.0, 1.2) \) & \( 0.0 \) & \( \mathcal{U}(1, 3) \) & \{walk, run\} \\
        Quadruped & \( \mathcal{U}(-0.1, 0.5) \) & \( \mathcal{U}(-0.3, 0.3) \) & \( \mathcal{U}(1, 3) \) & \{trot, jump\} \\
        \bottomrule
    \end{tabular}
    \label{tab:rollout_parameters}
\end{table}
We define a rollout for the biped to be a sequence of 5 sampled tuples $(v_x \times v_y \times T_{dur} \times b)$ whereas for the quadruped a rollout is defined as one tuple:
\begin{equation}
\label{eq:rollouts}
    \begin{aligned}
    & R_{\text{biped}} = \left( v_x \times v_y \times T_{dur} \times b \right)_{1:5},\\
    & R_{\text{quadruped}} = \left(v_x \times v_y \times T_{dur} \times b\right)
    \end{aligned}
\end{equation}

We define rollouts for the biped to have a sequence of sampled values in order to record expert demonstrations of gait switching behavior. Conversely, we do not record gait switching behavior for the quadruped because the rollouts only consist of one tuple of sampled values. This implies that gait switching is in-distribution for the biped but not the quadruped and we will leverage this later in the discussion of our results.
\subsubsection{Policy Parametrization}
All policies are parameterized as three-layer fully connected perceptrons with Relu activation functions. Hyperparameters are kept constant between the different policies that were compared. The four policies $\pi_{g_{1:4}}$ that are compared are each defined by their goal representation which is presented in Table \ref{tab:goal_representations}. In addition to the goal representation, the inputs to each policy also contain the current state of the robot which is the same across all policies.

\begin{figure*}
    \centering\includegraphics[width=1.0\textwidth]{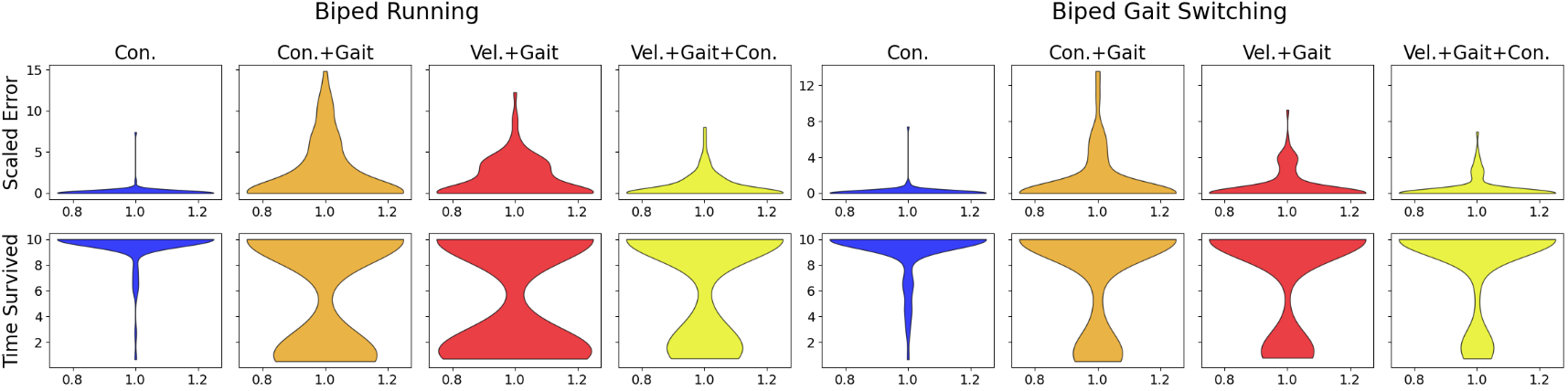}
    \caption{Biped in-distribution evaluations. Top: scaled error \eqref{metricp}, bottom: time survived in evaluation rollout, max 10 seconds. As we can see in the plots, \emph{Con.} policy has substantially smaller error and larger survival time compared to the other representations.}
    \label{fig:biped_in_distr}
\end{figure*}

\section{Results}\label{sec:result}

In this section, we present the results of the experiments comparing our contact-conditioned policy \emph{Con.} to the standard velocity-conditioned policy \emph{Vel.+Gait}. We also perform two ablations \emph{Vel.+Gait+Con.} and \emph{Con.+Gait} (see Table \ref{tab:goal_representations} for goal definitions). In particular, we trained the four policies to control the biped Bolt and the quadruped Go2. In the following we detail the setup for each case:
\begin{enumerate}
    \item \textbf{Biped}: the training data consisted of rollouts with walking, running and gait switching. We only collected locomotion data of the robot in the saggital direction in order to test the ability of the different policies to generalize to unseen velocities in the lateral direction.
    \item \textbf{Quadruped}: the training data included both sagittal as well as lateral velocities, however, this time we did not include episodes with gait switching. Including both sagittal and lateral velocities helps to see if the observations from the biped translate to more general cases. Furthermore, not including gait switching in the training data allows us to see if the policies are able to effectively transition between the different skills.
\end{enumerate}

Details on the episode sampling parameters are summarized in Table \ref{tab:rollout_parameters}. All policies were trained with the same expert demonstrations, with fixed hyperparameters, and for the same number of epochs. Our goal in this section is to answer the following question: \textbf{Which goal contains a richer representation for multi-gait policy learning?}

\textbf{Remark}: The policies encode multiple gaits in one neural network. All policies except the contact-conditioned policy (\emph{Con.}) receive information about which gait should be executed. The contact-conditioned policy can only infer the gait information from the contact goals.

\subsection{Evaluation Metrics}\label{metrics}

To compare the policies we will use two metrics, the first one is total survived time $T_{surv}$ in an evaluation episode rollout. 
% \begin{equation}
%     T_{surv}.
% \end{equation}
An episode terminates either at 10 seconds or earlier if any of the following conditions are violated:
\begin{itemize}
    \item Base height: \( 0.1 < h_{\text{base}} < 1 \) [m]
    \item Base Pitch: \( |\text{pitch}| \geq 30^\circ \)
    \item Base Roll: \( |\text{roll}| \geq 45^\circ \)
\end{itemize}

Furthermore to evaluate the performance of the policy we use the following scaled error $\epsilon$:

\begin{equation}\label{metricp}
    \epsilon = \frac{\left\| \bar{v}_{\text{expert}} - \bar{v}_{\text{policy}} \right\|}
    {\max\left(T_{\text{surv}} / T_{\text{max}}, \delta\right)}, \text{where } \delta = 0.1.
\end{equation}

The velocity error is scaled inversely with the survived percentage, ensuring that a balance between survived time and velocity error is captured.

\subsection*{1. Biped Experiments}
The biped was first evaluated in-distribution of its training data on a uniform grid of samples with saggital velocities $v_x \in [-1.0, 1.2]$. 100 episodes spanning this range were rolled out for walking, running and gait switching with a maximum duration of 10 seconds. 

The scaled error \eqref{metricp} and survival time (the time before failure) for in-distribution tests are plotted in Fig. \ref{fig:biped_in_distr}. As all walking policies saturated to the performance of the expert controller, we removed walking from this figure and only look at the more challenging cases of running and gait switching. As we can clearly see in Fig. \ref{fig:biped_in_distr}, the \emph{Con.} policy substantially outperform other representations, replicating the expert behavior with less failures. 

\begin{figure}
    \centerline{\includegraphics[scale=0.25]{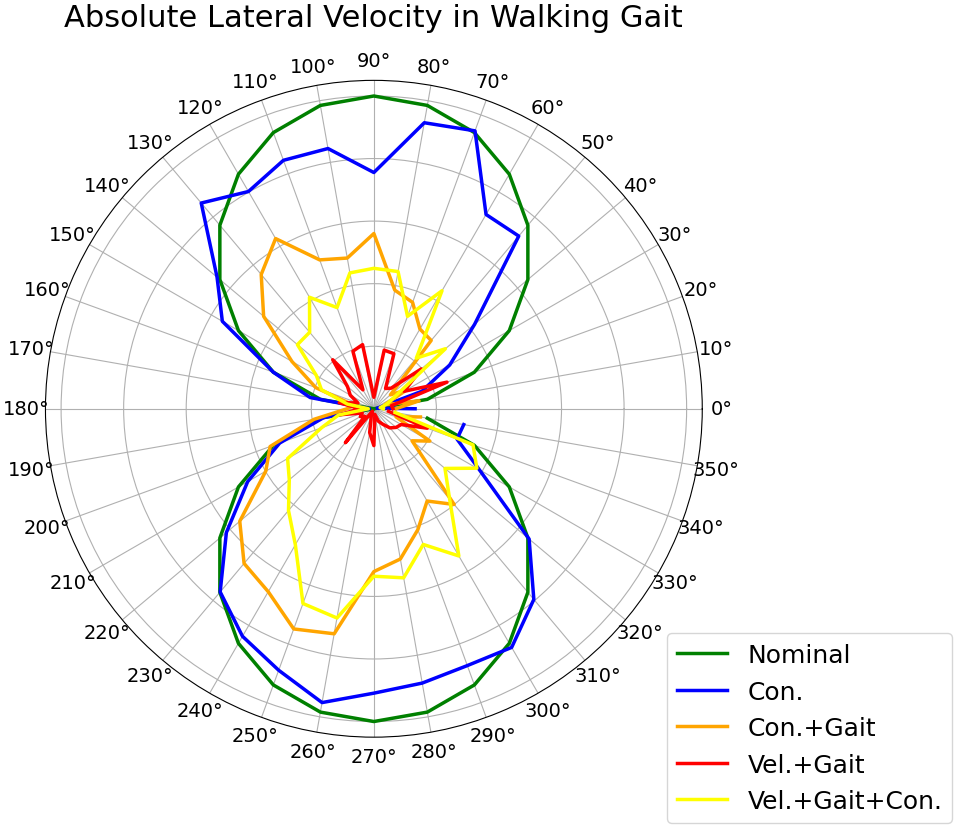}}
    \caption{Biped out-of-distribution evaluations. The robot was commanded to walk at 0.5 m/s in a dense grid of planar directions, while it was trained only on forward and backward motions. The rays represent the velocity aligned with the lateral plane of the robot. The green color denotes the nominal commanded lateral velocity. The \emph{con.} policy clearly outperforms other representations in tracking the nominal behavior in all directions.}
    \label{fig:biped_out_of_dist_polar}
\end{figure}

Next we performed an experiment testing the policies' ability to execute velocities not seen during training. This experiment consisted of the biped commanded to walk at 0.5 m/s in a dense grid of planar directions, namely at every 10 degrees of the unit circle while the robot was trained only on forward and backward motions. These results are visualized in Fig. \ref{fig:biped_out_of_dist_polar}, where the rays represent the velocity aligned with the lateral plane of the robot. The green color denotes the nominal commanded lateral velocity, hence the closer the results are to the green the better the performance. As it can be seen in Fig. \ref{fig:biped_out_of_dist_polar}, the contact-conditioned (blue) policy was able to most closely follow the desired lateral velocity (green) while the \emph{Vel.+Con.} (Yellow) and \emph{Con.+Gait} (orange) policies showed moderate ability. As expected, the velocity-conditioned (red) policy was not able to track velocities not seen in training, demonstrating how overfitting to high-level skills can hamper generalization. Basically policies containing contact information were able to execute velocities in the lateral direction, which supports the hypothesis that conditioning on contact information improves out-of-distribution generalization.
% Although these policies were trained on the same data, the contact goal representation captures an invariant representation of the locomotion behavior, allowing it to follow velocities it was not trained on.
\begin{figure*}[t]
    \centering
    \includegraphics[width=1.0\textwidth]{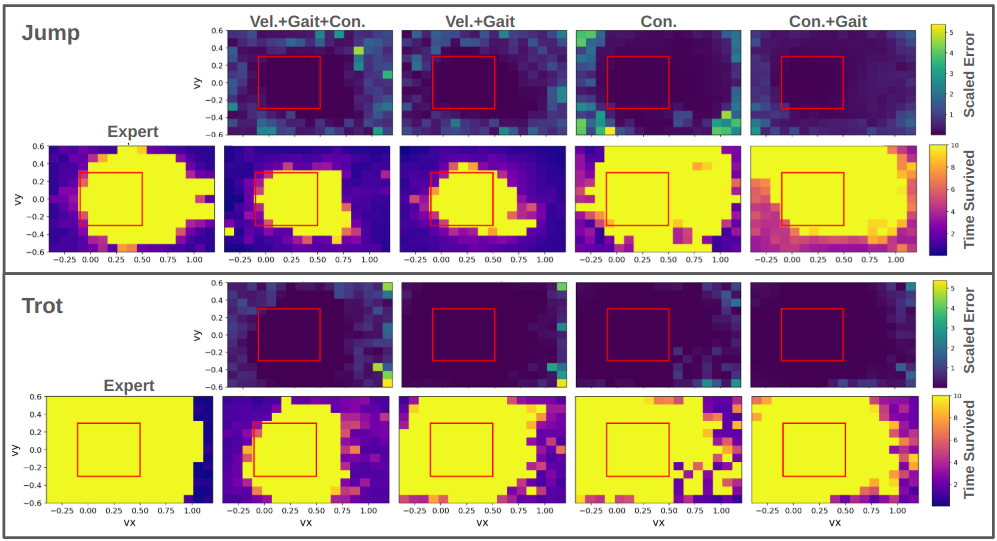}
    \caption{Quadruped velocity tracking evaluations. Policies were tested on a dense grid of x and y velocities for jump (top) and trot (bottom). Inside the red box is the velocity range present in the training data. Top(blue-green): scaled error \eqref{metricp}. Bottom(yellow-purple): time survived in evaluation rollout, max 10 seconds. While all policies perform similarly inside the range of velocities they trained on, contact-conditioned policy outperforms others for the out-of-distribution velocities.}
    \label{fig:quadruped_unified_data}
\end{figure*}
\subsection*{2. Quadruped Experiments}

Similar to the biped case, we evaluated the quadruped experiments both in distribution as well as out of distribution of its training data. To do so, we tested policies on a uniform grid of 221 episodes for the trotting and the jumping gait separately and with velocities in the range $v_x \in [-0.4, 1.2], \quad v_y \in [-0.6, 0.6]$. The scaled error \eqref{metricp} and survival time in this case are visualized in Fig. \ref{fig:quadruped_unified_data}. The upper plot for each gait shows via the blue-green heatmaps the scaled error $\epsilon$ with a high number (green) implying worse performance. The purple-yellow heatmaps show the survivability $T_{surv}$ with yellow implying long survivability and purple representing early termination. Furthermore, the cells within the red boxes represent the velocity range present in the training data.
As we can see in these plots, inside the distribution of the training data in the red boxes, the results are very similar (slightly better for the contact-conditioned policies). This is in contrast to the biped case, where even inside the distribution of the training data there were stark differences. We believe that this is likely due to the fact that the quadrupeds are much more stable than bipeds and the motions are less sensitive to the errors the control policy makes.
However, when looking at the results outside of the red boxes, namely the out of distribution data, it can clearly be seen in Fig. \ref{fig:quadruped_unified_data} that the survivability regions of the \emph{Con.} and \emph{Con.+Gait} policies are larger compared to the \emph{Vel.+Gait} and \emph{Vel.+Gait+Con.} policies. This result is consistent with the result of the biped and further strengthens our claim that the contact goal representation captures an invariant representation of the locomotion behavior, allowing it to follow velocities it was not trained on. 
Interestingly, the \emph{Vel.+Gait+Con.} policy performs the worst outside of the distribution of the training data. This is an interesting observation as it shows that providing more information to the policy does not necessarily improve the performance and might lead to overfitting to the redundant information. 
% Within the red boxes, this pattern is not visible, therefore we conclude that with this representation the learning process is susceptible to overfitting. We suspect that the contact and velocity goals may be conflicting with eachother, hampering the performance of the policy.

\begin{figure*}[t]
    \centering
    \includegraphics[width=1.0\textwidth]{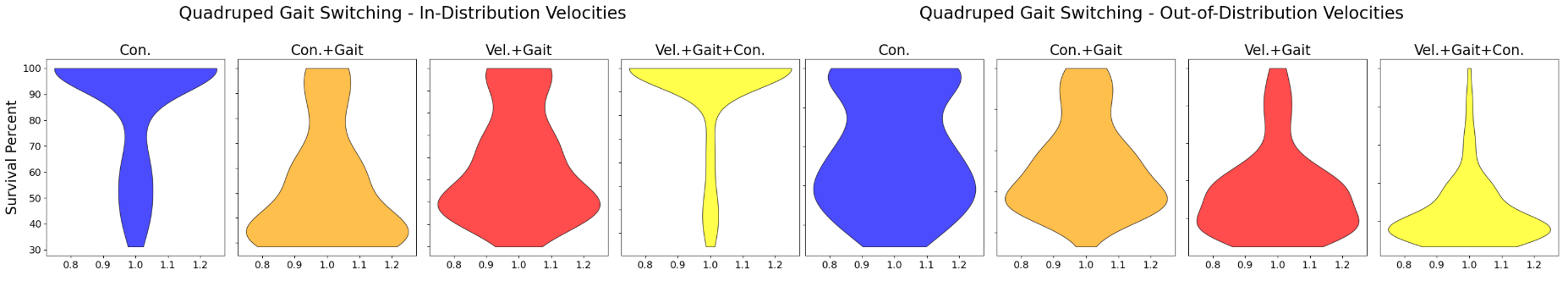}
    \caption{Quadruped gait switching evaluations. y-axis: percent of time survived in evaluation rollout. Left: in-distribution velocities. Right: out-of-distribution velocities. While \emph{vel.+con+gait} shows competitive results compared to \emph{con.} inside the training distribution, its performance drops outside of the training distribution and the \emph{con.} policy shows the best survivability.}
    \label{fig:quadruped_gait_switching}
\end{figure*}

In the final experiment, we test the gait switching capability in episodes with randomized switching times and randomized gaits. The results are illustrated in Fig. \ref{fig:quadruped_gait_switching}, for both inside (left) and outside (right) of the distribution of the training velocities. It is important to note that the policies in this case were not trained on gait switching data, therefore we consider this as out-of-distribution. Surprisingly, the \emph{Vel.+Gait+Con.} performed the best gait switching alongside \emph{Con.} on the in-distribution velocities. This was unexpected, because in the other experiments \emph{Vel.+Gait+Con.} performed poorly. Perhaps we overestimated the difficulty of gait switching while executing in-distribution velocities, however when looking at the results for the out-of-distribution velocities on the right hand side of Fig. \ref{fig:quadruped_gait_switching}, we see that these results align with our previous experiments. We can see an extreme drop in survivability for the \emph{Vel.+Gait+Con.}, mirroring its out-of-distribution performance in the previous experiment. 

In contrast to overfitting and similarly to the previous experiments, the \emph{Con.} followed by the \emph{Con.+Gait} policies show the best out-of-distribution results for the gait switching on the quadruped platform, further reinforcing our claim that the contact-conditioned \emph{Con.} policy is a suitable goal representation for a multi-gait policy.

%% End Notes
\section{Conclusion}\label{sec:conclusion}
In this paper, we evaluated the contact mode goal representation for multi-gait legged locomotion. Contrary to traditional conditioning on the desired average velocity, we proposed to represent the goal as a set of desired future contact switches. Through an extensive set of simulation experiments on the walking and running of a biped robot as well as trotting and jumping of a quadruped, we have shown that representing desired behavior as a set of next contact switches can improve the performance and robustness in both in-distribution and out-of-distribution scenarios. In particular, our results showed the promise of contact-conditioned policies for out-of-distribution generalization.

In the future, we plan to go beyond cyclic gaits and train a generalist policy that can realize any desired contact sequences from a contact planner. Furthermore, we are interested in testing the hypothesis on manipulation and loco-manipulation tasks, as contact is central for both locomotion and object manipulation problems. Finally, real-world experiments are foreseeable in the near future.

%%%%%%%%%%%%%%%%%%%%%%%%%%%%%%%%%%%%%%%%%%%%%%%%%%%%%%%%%%%%%%%%%%%%%%%%%%%%%%%%
\bibliography{master} 
\bibliographystyle{ieeetr}

\end{document}